\documentclass[journal]{IEEEtran}
\ifCLASSINFOpdf
\else
\fi
\usepackage{comment}
\usepackage{algpseudocode}
\usepackage{xcolor}
\usepackage{multirow}
\usepackage{multicol}
\usepackage{times}
\usepackage{epsfig}
\usepackage{amsmath}
\usepackage{amssymb}

\usepackage{siunitx}

\usepackage{graphicx}
\usepackage{booktabs}
\usepackage{algorithm}
\usepackage{cite}
\usepackage[font=small,labelfont=bf]{caption}
\usepackage{subcaption}
\usepackage{caption, subcaption}
\usepackage[pagebackref=true,breaklinks=true,colorlinks,bookmarks=false,citecolor=green]{hyperref}
\newcommand{\R}[1]{\textcolor{red}{#1}}
\newcommand{\G}[1]{\textcolor{green}{#1}}

\newcommand{\figref}[1]{Fig.~\ref{#1}}

\newcommand{\tabref}[1]{Tab.~\ref{#1}}

\newcommand\xsqcomments[1]{\textcolor{green}{#1}}
\newcommand{\GRE}[1]{\textcolor{green}{#1}}
\newcommand{\RED}[1]{\textcolor{red}{#1}}

\begin{document}
\sloppy 
%
\title{Multi-Sem Fusion: Multimodal Semantic Fusion for 3D Object Detection}
%
%
%

\author{Shaoqing Xu, Fang Li, Ziying Song, Jin Fang, Sifen Wang,  Zhi-Xin Yang, \textit{Member, IEEE}
\thanks{This work was funded in part by the Science and Technology Development Fund, Macau SAR (Grant no. 0018/2019/AKP, SKL-IOTSC(UM)-2021-2023 and 0059/2021/AFJ), in part by the Guangdong Science and Technology Department, China (Grant no. 2020B1515130001), in part by the University of Macau (Grant No.: MYRG2020-00253-FST and MYRG2022-00059-FST), and in part by the Zhuhai UM Research Institute (Grant No.: HF-011-2021). \emph{(Corresponding author: Zhi-Xin Yang).}
}
\thanks{Shaoqing Xu and Zhi-Xin Yang are with the State Key Laboratory of Internet of Things for Smart City and Department of Electromechanical Engineering, University
of Macau, Macau 999078, China (e-mail: xushaoqing26@gmail.com, zxyang@um.edu.mo)
}

\thanks{Fang Li is with School of Mechanical Engineering, Beijing Institute of Technology. (e-mail:a319457899@163.com)}

\thanks{Ziying Song is with School of Computer and Information Technology, Beijing Key Lab of Traffic Data Analysis and Mining, Beijing Jiaotong University, Beijing 100044, China (e-mail: 22110110@bjtu.edu.cn)
}

\thanks{Sifen Wang is with School of Transportation Science and Engineering, Beihang University, Beijing 100083, China (e-mail: sfwang@buaa.edu.cn)}
\thanks{Fang Jin is with Inceptio, email: jin.fang@inceptio.ai}

}


%
%

 
\markboth{Journal of \LaTeX\ Class Files,~Vol.~14, No.~8, August~2015}%
{Shell \MakeLowercase{\textit{et al.}}: Bare Demo of IEEEtran.cls for IEEE Journals}
%
\maketitle
\begin{abstract}
LiDAR and camera fusion techniques are promising for achieving 3D object detection in autonomous driving. Most multi-modal 3D object detection frameworks integrate semantic knowledge from 2D images into 3D LiDAR point clouds to enhance detection accuracy. Nevertheless, the restricted resolution of 2D feature maps impedes accurate re-projection and often induces a pronounced boundary-blurring effect, which is primarily attributed to erroneous semantic segmentation. To well handle this limitation, we propose a general multi-modal fusion framework \textit{Multi-Sem Fusion (MSF)} to fuse the semantic information from both the 2D image and 3D points scene parsing results. 
Specifically, we employ 2D/3D semantic segmentation methods to generate the {parsing results} for 2D images and 3D point clouds. The 2D semantic information is further re-projected into the 3D point clouds with calibration parameters. To handle the misalignment between the 2D and 3D parsing results, we propose an Adaptive Attention-based Fusion (AAF) module to fuse them by learning an adaptive fusion score. Then the point cloud with the fused semantic label is sent to the following 3D object detectors. Furthermore, we propose a Deep Feature Fusion (DFF) module to aggregate deep features at different levels to boost the final detection performance. The effectiveness of the framework has been verified on two public large-scale 3D object detection benchmarks by comparing them with different baselines. The experimental results show that the proposed fusion strategies can significantly improve the detection performance compared to the methods using only point clouds and the methods using only 2D semantic information. Most importantly, the proposed approach significantly outperforms other approaches and sets state-of-the-art results on the nuScenes testing benchmark.
\end{abstract}


%
\IEEEpeerreviewmaketitle

%
%
%
%

 



\section{Introduction}\label{sec:intro}
\IEEEPARstart{V}{sion-based} perception tasks, like \textbf{3D Object Detection}, semantic segmentation, lane detection, has been extensively studied with the development of Autonomous Driving(AD) and intelligent transportation system \cite{wang2022pursuing,HANet}. Among them, LiDAR-based 3D Object detection is the mainstream of current research. The depth information can be easily got from LiDAR sensors to localize objects. However, the texture and color information has been totally lost due to the sparse scanning. Therefore, False Positive (FP) detection and wrong categories classification often arises for LiDAR-based object detection frameworks.

\begin{figure}[t!]
	\centering
	\includegraphics[width=0.45\textwidth]{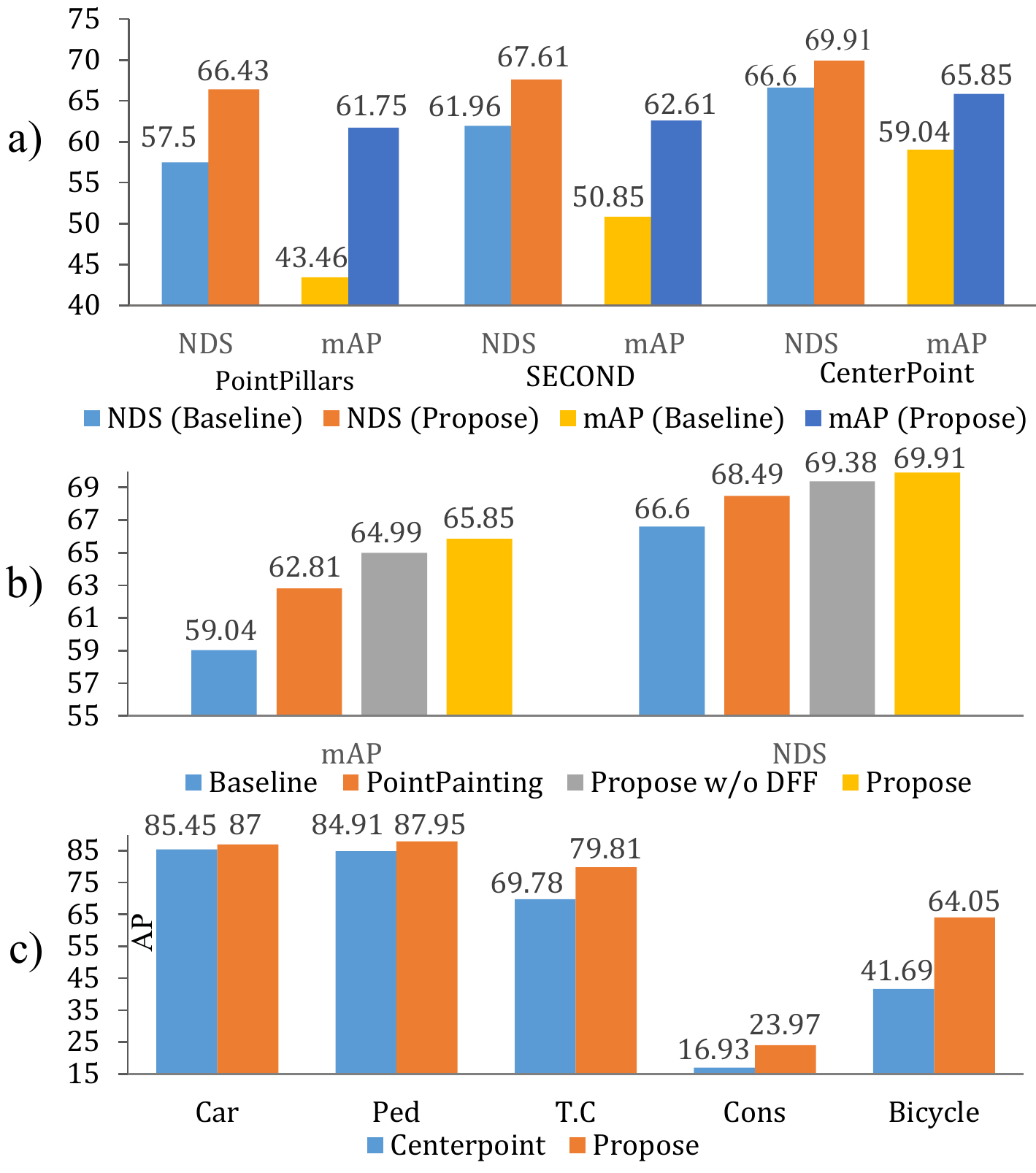}
	\centering
	\caption{The proposed \textit{Multi-Sem Fusion (MSF)} is a general multi-modal fusion framework which can be employed for different 3D object detectors. a) illustrate the improvements on three different baselines. b) gives the performance of \textit{CenterPoint} \cite{yin2021center} with the proposed modules on the public nuScenes benchmark. c) gives the improvements on different categories respectively. In addition, ``w/o'' represent ``without'' in short.}  
	\label{Fig:head_figure}
\end{figure}

On the other hand, images can provide details texture and color information while the depth information has been totally lost. 
The fusion method with various sensor data is an encouraging way for boosting the perception performance of AD. Generally, multi-modal fusion approaches in object detection tasks can be divided into early fusion-based \cite{dou2019seg, vora2020pointpainting}, deep fusion-based \cite{zhang2020maff, xu2018pointfusion, tian2020adaptive} and late fusion-based approaches \cite{pang2020clocs}. Early fusion-based approaches aim to create a new type of data by directly combining the raw data before inputting it into the detection framework. Typically, such methods require pixel-level correspondence between different sensor data types. Different from the early fusion-based methods, late fusion-based approaches {fuse the detection results at the bounding box level. While} deep fusion-based methods usually extract the features with different types of deep neural networks first and then fuse them at the features level.  Currently, most multi-modal 3D object detection frameworks leverage semantic information from 2D images to improve detection accuracy in 3D LiDAR point clouds. For instance, \textit{PointPainting} \cite{vora2020pointpainting} , a classic early fusion-based approach, takes both the point cloud and 2D image semantic predictions as input and outputs detection results, which can be utilized with any LiDAR-based 3D object detector implemented using either point cloud or voxel-based frameworks. 

\begin{figure}[t]
	\centering
	\includegraphics[width=0.45\textwidth]{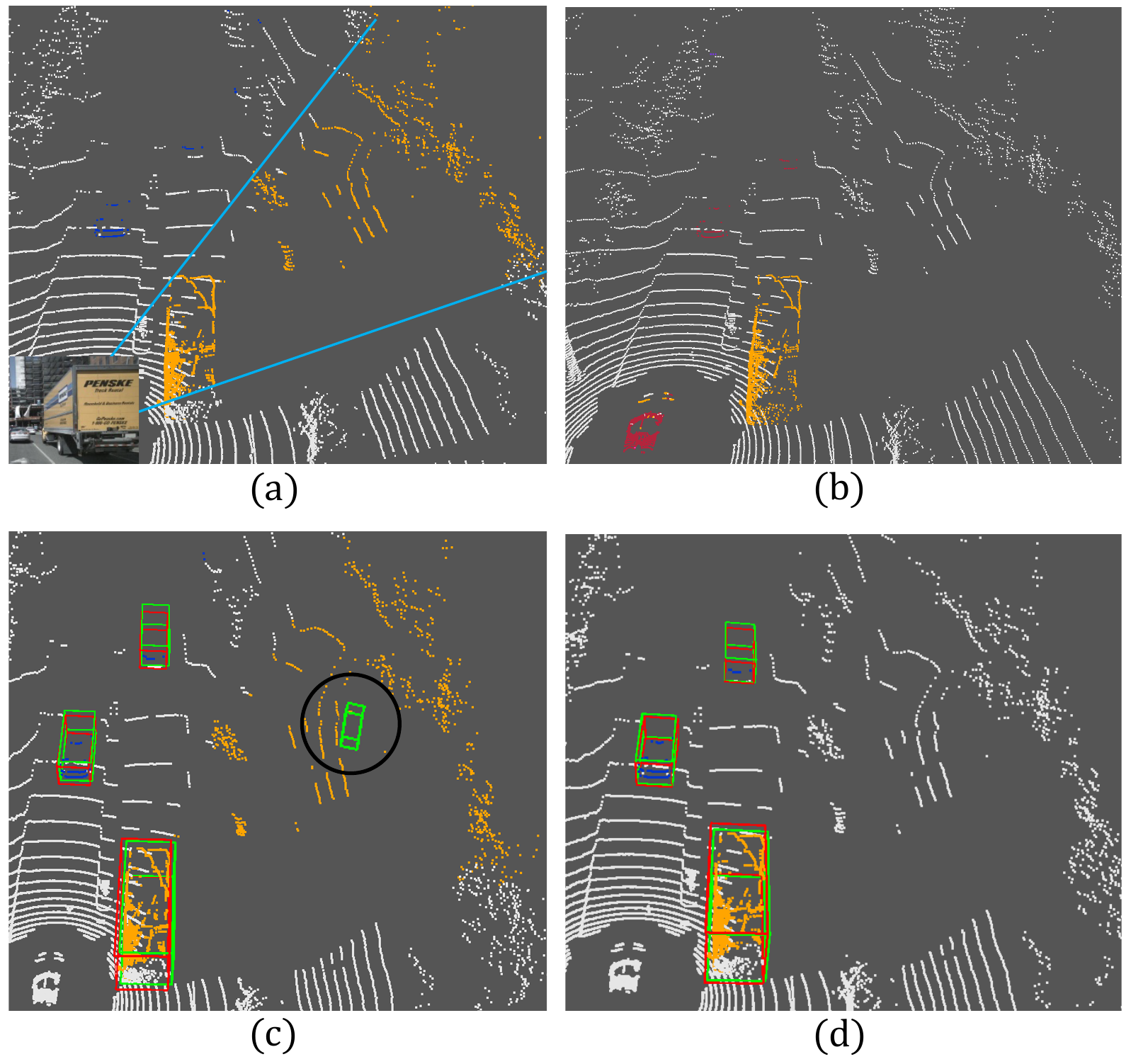}
	\centering
	\caption{(a) shows the point cloud with the 2D segmentation results, where the frustum within the blue line highlight indicates misclassified areas due to the blurring effect at the object's boundary; (b) displays the 3D segmentation results, where misclassified points are colored in red; (c) and (d) show the results based on 2D painted point cloud (with an obvious False Positive (FP) Detection) and the proposed \textit{Multi-Sem Fusion (MSF)} framework respectively.} 
	\label{Fig:2DSeg_in_3d}
\end{figure}

However, the {blurring effect at object's boundary} often happens inevitably in image-based semantic segmentation methods. This effect becomes much worse when reprojecting the 2D semantic projections into the 3D point clouds. An example of this effect has been shown in Fig. \ref{Fig:2DSeg_in_3d}. Taking the big truck at the left bottom of the sub-fig, an instance of this issue is illustrated in Fig. \ref{Fig:2DSeg_in_3d}-(a), the significant frustum area in the blue line highlight of the background (i.e., orange points) incorrectly classified as foreground due to inaccurate projection. Additionally, the re-projection of 3D points onto 2D image pixels is not a one-to-one process because of the digital quantization and many-to-one projection issues. Notably, segmentation results from the 3D point clouds (as seen in Fig. \ref{Fig:2DSeg_in_3d}-(b)) outperform the 2D image at obstacle boundaries. However, the category classification from 3D point clouds often yields worse results (as demonstrated by points in red) when compared to the 2D image, primarily due to the loss of detailed texture information in point clouds.

The effectiveness of utilizing 2D image semantic information for 3D object detection has been demonstrated, as exemplified by PointPainting\cite{vora2020pointpainting}, despite the presence of some semantic errors. However, a natural question arises: can the final detection performance be further improved by fusing both the 2D and 3D semantic results in an effective way? To address this question, we introduce a general multi-modal fusion framework \textit{Multi-Seg Fusion} which fusing the multi-modal data at the semantic level to improve the final 3D object detection performance. First of all, {we obtain the 2D/3D semantic information throughout 2D/3D parsing approaches by using images and raw point clouds. Then, each point has two types of semantic information after projecting point clouds onto 2D semantic images based on the intrinsic and extrinsic calibration parameters.} However, the semantic results conflict usually happens for a certain point, rather than concatenating the two types of information directly, we propose an AAF strategy to fuse different types of semantic information in point or voxel-level adaptive manner. It's achieved by the learned context features with the self-attention mechanism. Specifically, attention scores are learned for each point or voxel to balance the importance of the two different semantic results.
 
Furthermore, {in order to detect  objects with different sizes in an efficient way}, a DFF module is proposed here to fuse the features {at multi-scale receptive fields and a channel attention network to gather related channel information in feature map}. Then the fused features are passed for the following classification and regression heads to generate the final detection. The results on nuScenes dataset are given in \figref{Fig:head_figure} (a), and we can obviously find that the proposed modules can robustly boost the detection performance on different baselines. {The results in \figref{Fig:head_figure}(b) illustrates the contribution of each module} and it can also be observed that the detection results can be consistently improved by adding more modules gradually. 
\figref{Fig:head_figure} (c) shows the improvements on different categories and we can easily find that all the classes have been improved and ``Bicycle'' gets the most improvement (e.g., 22.3 points) on the nuScenes dataset.
 

   Compared to the \textit{FusionPainting} \cite{xu2021fusionpainting},  We enhanced it with new relevant state-of-the-art methods, theoretical developments, and experimental results. This work significantly improves the 3D object detection accuracy hugely on the nuScenes 3D object detection benchmark. Moreover, we also evaluate the proposed framework on the KITTT 3D object detection dataset, and the experimental results on both public datasets demonstrate the superiority of our framework. In generally, this work can be characterized by the following contributions: 
\begin{enumerate}
 \item 
 The proposed \textit{Multi-Sem Fusion} framework offers a general approach for multi-modal fusion at semantic level, which improves 3D object detection performance by integrating multi-modal data.
   \item Rather than combining different semantic results directly, an adaptive attention-based fusion module 
   is introduced, which learns fusion attention scores instead of directly combining the results.
   \item Furthermore, a deep feature fusion module is proposed to fuse deep features at different levels to better detect objects of various sizes. This is achieved by fusing deep features extracted from different layers of the network and channel attention technology.
   \item The proposed fusion framework for 3D object detection is evaluated on two public benchmarks, demonstrating its superiority and achieving state-of-the-art (SOTA) results on the KITTI and nuScenes dataset. Taking the proposed framework as the baseline, we also won the champion in the fourth nuScenes object detection challenge at ICRA Workshop.
\end{enumerate}



\section{Related Work}\label{sec:related_work}

\subsection{{ Single-sensor 3D Object Detection}}
Typically, the classification of single-sensor 3D Object Detection methods can be categorized into two groups: LiDAR-based and image-based approache, which take LiDAR point cloud and image-captured data as inputs, respectively. 

\textbf{LiDAR-only.} The existing LIDAR-{based} 3D object detection methods {can be generally categorized into three main groups as} projection-based~\cite{ku2018joint,luo2018fast,yang2018pixor,liang2021rangeioudet,hu2020you}, voxel-based~\cite{zhou2018voxelnet,yan2018second,kuang2020voxel, yin2020lidar,zhou2019iou,vpnet,wu2022casa} and point-based~\cite{zhou2020joint,lang2019pointpillars,qi2018frustum}. 
RangeRCNN \cite{liang2021rangeioudet} proposed a 3D range {image-based} 3D object detection framework {where} the anchors could be generated on the BEV (bird’s-eye-view) map. VoxelNet\cite{zhou2018voxelnet} is a voxel-based framework that utilizes a VFE layer to extract point-level features for each voxel. SECOND \cite{yan2018second} proposes a sparse convolution operation to replace heavy 3D convolutions for faster inference. CenterPoint\cite{yin2021center} presents an anchor-free approach for 3D object detection and achieves state-of-the-art performance on the nuScenes benchmark. PointPillars\cite{lang2019pointpillars} divides points into vertical columns and extracts features with PointNet \cite{qi2017pointnet}, allowing for use in 2D object detection pipelines for 3D object detection. PointRCNN\cite{shi2019pointrcnn} extracts features directly from raw point clouds and generates proposals from foreground points. PV-RCNN \cite{shi2020pv} combines voxel and point representations and employs both 3D voxel-based CNNs and PointNet-based networks for improved discriminative feature learning.

\textbf{Camera-only.} In previous image-based 3D object detection methods, features are extracted by constructing a network from either single or multiple images for predicting 3D bounding boxes. Some monocular-based approaches\cite{Simonelli_Bulò_Porzi_Lopez-Antequera_Kontschieder_2019,Wang_Zhu_Pang_Lin_2021} attempt to regress and predict 3D boxes directly from a single image, while others\cite{Wang_Chao_Garg_Hariharan_Campbell_Weinberger_2019,You_Wang_Chao_Garg_Pleiss_Hariharan_Campbell_Weinberger_2019} suggest constructing intermediate-level representations and performing detection on top of them. Depth estimation \cite{Ma_Kundu_Zhu_Berneshawi_Ma_Fidler_Urtasun_2015,Reading_Harakeh_Chae_Waslander_2021,Shi_Chen_Kim_2020} has also been used to enhance 3D detection ability due to its necessity in the process. Another approaches \cite{Chen_Han_Xu_Su,Chen_Liu_Shen_Jia_2020,Yao_Luo_Li_Fang_Quan_2018} for obtaining relatively accurate depth use stereo or multi-view images to create 3D geometry volumes for object detection. However, although depth estimated from multi-view images is better than that from a single image, it still lags behind the accuracy achieved with point clouds.

\begin{figure*}[ht!]
	\centering
	\includegraphics[width=1\textwidth]{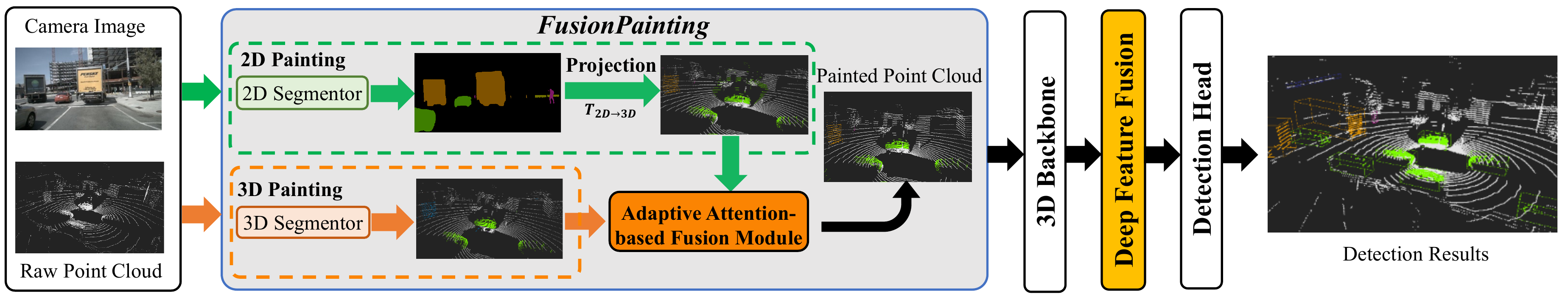}
	\centering
	\caption{Overview of the proposed \textit{Multi-Sem Fusion} framework. We first obtain the semantic information from both input point clouds and 2D images with 2D and 3D parsing approaches. The semantic information from the two types of data is then fused at the semantic level using the proposed AAF module. Furthermore, a DFF module is also proposed to fuse the deep features at different spatial levels to boost the detection for accurate kinds of size object. Finally, fused features are sent to the detection heads for producing the final detection results.
}
	\label{Fig:PointPainting}
\end{figure*}


\subsection{{Multi-sensors Fusion-based 3D Object Detection}}
{LiDAR can provide reliable depth information while the textures information has been lost while camera sensor is just the opposite. Multi-modal 3D object detection utilizes the advantages of both structure information in point cloud and textures information in image. According to \cite{nie2020multimodality}, multi-modal fusion approaches can be categorized into model-based \cite{muresan2020stabilization} and data-based methods based on the way of fusing the sensor data. Generally, model-based methods have been widely utilized for tracking \cite{muresan2019multi} while data-based approaches are more commonly applied to environment perception tasks like object detection \cite{du2018general,pang2020clocs,chen2017multi}. F-PointNet \cite{qi2018frustum}, PointFusion \cite{xu2018pointfusion} and \cite{du2018general} generate object proposal in the 2D image first and then fuse the image features and point cloud for 3D BBox generation.  
PointPainting \cite{vora2020pointpainting} is an approach to enhance the performance of 3D point cloud object detectors by fusing semantic information from 2D images into the point cloud. In order to obtain parsing results, any SOTA approach can be used. AVOD \cite{ku2018joint}, which is a typical late fusion framework, generates features that can be shared by RPN (region proposal network) and second stage refinement network. 
Multi-task fusion \cite{liang2019multi} has been proven to be effective in various tasks, such as jointing semantic segmentation with 3D object detection tasks as demonstrated in \cite{zhou2020joint}. Furthermore, Radar and HDMaps are also employed for improving the detection accuracy.  In \cite{yang2018hdnet} and \cite{fang2021mapfusion}, the HDMaps have been utilized as a valuable source of prior information for detecting moving objects in AD scenarios. Using the frustum-based association method, CenterFusion \cite{nabati2021centerfusion} focuses on the fusion of radar and camera sensors. It associates the radar detections with objects in the image and creates radar-based feature maps to supplement the image features through a middle-fusion approach.

\section{Multi-Sem Fusion Framework} \label{sec:method}
As shown in \figref{Fig:PointPainting} provides an overview of the \textit{Multi-Sem Fusion} framework which is designed to fully leverage the information obtained from a variety of sensors, we advocate fusing them at two different levels. First, the two types of information are early fused with the \textit{PointPainting} technique by painting the point cloud with both the 2D and 3D semantic parsing results. To handle the {inaccurate segmentation} results, an AAF module is proposed to learn an attention score for different sensors {for the following fusion purpose}. By taking the points {together with fused semantic information} as inputs, deep features can be extracted from the backbone network. By considering that different size object requires different levels features, a novel DFF module is proposed to enrich features from the backbone with different levels for exploring the global context information and the local spatial details in a deep way. As illustrated in \figref{Fig:PointPainting}, Our framework comprises three primary models: a multi-modal semantic segmentation module, an Adaptive-attention fusion (AAF) module, and a deep feature fusion (DFF) module. To obtain the semantic segmentation from both RGB images and LiDAR point clouds, we can leverage any existing 2D and 3D scene parsing approaches. Then the 2D and 3D semantic information are fused by the AAF module. Bypassing the points {together with fused semantic labels} into the backbone network, the DFF module is used to {further improve the results by aggregating the features information within various receptive fields and a channel attention module.}
    

\begin{figure*}[ht]
    \centering
	\includegraphics[width=1\textwidth]{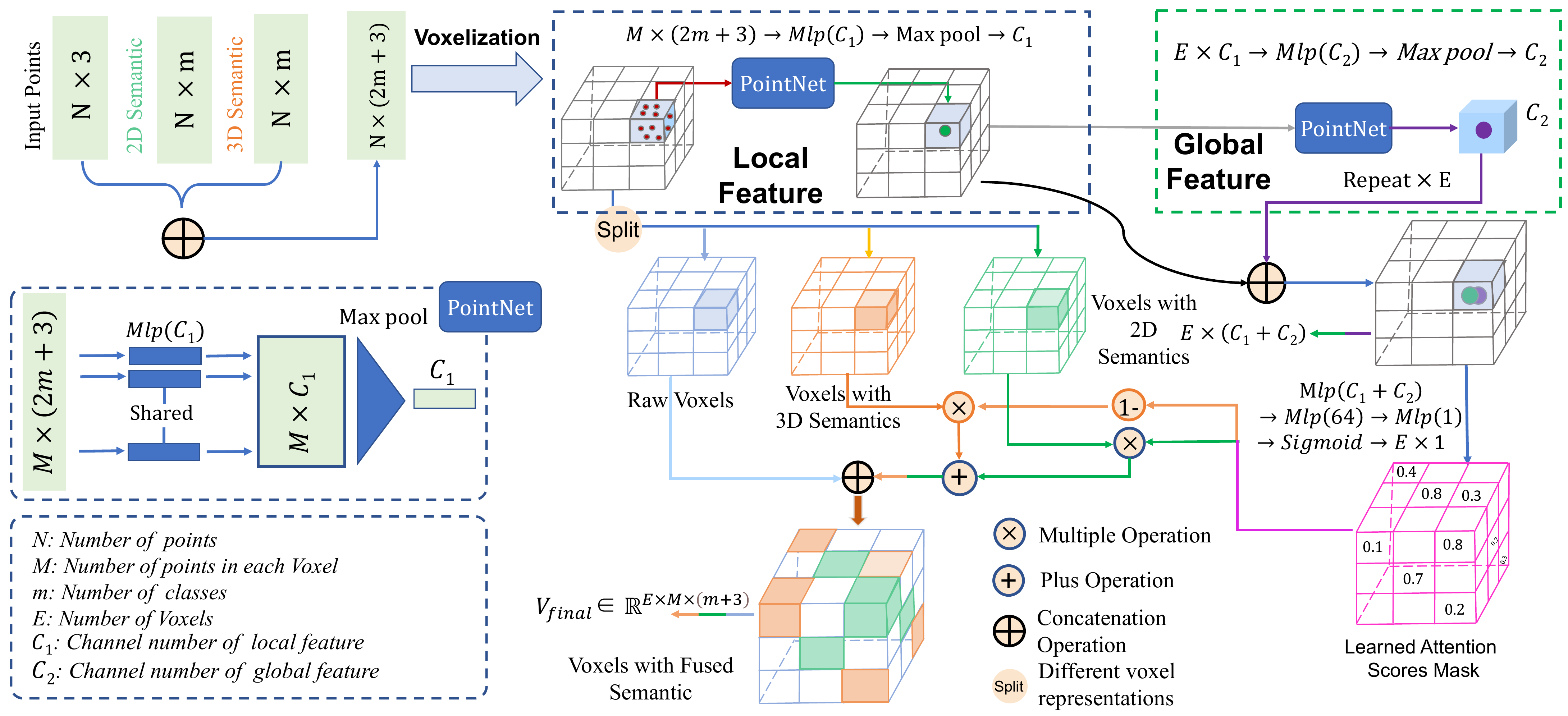}
	\caption{The architecture of the proposed AAF module for 2D/3D semantic fusion. The proposed framework leverages the input points and 2D/3D semantic results to learn attention scores throughout an adaptive attention network, which are then used to paint the raw points or voxel with adaptive 2D/3D semantic labels}
	\label{Fig:attention_module}
\end{figure*}
\subsection{2D/3D Semantic Parsing}

\textit{{2D Image Parsing.}} By incorporating 2D images into our approach, we can leverage their rich texture and color information to complement the analysis of 3D point clouds. For acquiring 2D semantic labels, a modern semantic segmentation method is employed here for generating pixel-wise segmentation results. More importantly, the proposed framework is allowing for the use of a wide range of state-of-the-art segmentation approaches without requiring modification. This agnostic approach enables researchers and practitioners to more easily leverage their preferred segmentation models and techniques (e.g., \cite{choudhury2018segmentation,zhao2017pyramid,chen2019hybrid,shelhamer2017fully,Howard_2019_ICCV}, etc). {We employ Deeplabv3+ \cite{choudhury2018segmentation} for generating the semantic results here.} The network takes 2D images as input and produces pixel-wise semantic classes scores for both the foreground and background categories. {Assuming that the obtained semantic map is $S \in \mathbb{R}^{w\times h \times m}$, where $(w, h)$ is the input image size and $m$ is the number of category. By employing the intrinsic and extrinsic matrices, the 2D semantic information can be easily re-projected into the 3D point cloud. Specifically, by assuming that} the parameter of the intrinsic matrix is $\mathbf{K} \in \mathbb{R}^{3\times 3}$, extrinsic matrix $\mathbf{M}\in \mathbb{R}^{3\times 4}$ and the original 3D points clouds is $\mathbf{P}\in \mathbb{R}^{N\times 3}$, we can derive the projection of the LiDAR 3D point cloud onto the camera as Eq.\eqref{eq.projection} shown:
\begin{equation} \label{eq.projection}
	 {\mathbf{P}}^{'} = \text{Proj}(\mathbf{K}, \mathbf{M}, \mathbf{P}),
\end{equation} 
where $\mathbf{P}^{'}$ represent the LIDAR point in camera coordinates and ``Proj'' denote the projection process. {By this projection, we can assign the semantic segmentation results obtained from the 2D image to their corresponding 3D points which is denoted by $\mathbf{P}_{2D}\in \mathbb{R}^{N\times m}$.



\textit{{3D Point Cloud Parsing.}} As we have mentioned above, {parsing results from the point clouds can well overcome the boundary blur influence while keeping the distance information. Similar to the 2D image segmentation, any SOTA 3D parsing approach can be employed here \cite{zhang2020polarnet,cheng20212,xu2021rpvnet}. We employ the Cylinder3D \cite{cong2021input} for generating the semantic results because of its impressive performance on the AD scenario.}
{More importantly, the ground truth point-wise semantic annotations can be generated from the 3D object bounding boxes roughly as \cite{shi2019pointrcnn} while any extra semantic annotations are not necessary.} Specifically, for assigning semantic labels to the 3D points, we directly assign class labels to the points inside a 3D bounding box for foreground instances, while considering points outside all the 3D bounding boxes as the background. This approach enables our proposed framework to work directly on 3D detection benchmarks without requiring additional point-wise semantic annotations. After training the network, we will obtain the parsing results which is denoted by $\mathbf{P}_{3D}\in \mathbb{R}^{N\times m}$.

\subsection{Adaptive Attention-based 2D/3D Semantic Fusion} \label{subsec:shallow_fusion}
{As mentioned in previous work PointPainting \cite{vora2020pointpainting}, 2D semantic segmentation network have achieved impressive performance,} however, the {blurring effect at the shape boundary} is also serious due to the restricted resolution of the feature map (e.g., $\frac{1}{4}$ of the original image size). Therefore, {the point clouds with 2D semantic segmentation} usually has misclassified regions around the objects' boundary. For example, the frustum region is illustrated in the sub-fig.~\ref{Fig:2DSeg_in_3d} (a) behind the big truck {has been totally misclassified}. On the contrary, the {parsing results} from the 3D point clouds usually perform a clear and accurate object boundary without blurring effect e.g., sub-fig.~\ref{Fig:2DSeg_in_3d}(b). However, the disadvantages of the 3D segmentor are also obvious. One drawback is that without the color and texture information, the 3D segmentor is difficult to distinguish these categories with similar shapes from the point cloud-only. In order to boost advantages while suppressing disadvantages, an AAF module has been proposed to adaptively combine the 2D/3D semantic segmentation results. Then the Optimized semantic information is ready for the following 3D object detectors backbone to further extract the enhanced feature to improve the final object detection accuracy results.


\textbf{AAF Module.} The detailed architecture of the proposed AAF module is depicted in Fig.~\ref{Fig:attention_module}. The input point clouds is defining as a set of points $\{\mathbf{P}_i, i=1,2,3...N\}$, with each point $\mathbf{P}_i$ containing $(x, y, z)$ coordinates and other optional information such as intensity. In the following context, we'll focus solely on the coordinates that meaning only the coordinates $(x, y, z)$ are considered as input data. Our objective is to develop an efficient strategy for integrating semantic information from both 2D images and 3D point clouds.}. Here, we propose a novel approach that employ adaptive attention fusion(AAF) module to learn each cell attention score in voxel level or point level to adaptively combine the two types of semantic results. Specifically, the module 
begins by concatenating the point clouds coordinate attributes $(x, y, z)$ and 2D/3D semantic segmentation labels to obtain a fused point clouds with the shape of $N\times (2m+3)$. In order to reduce memory consumption during the fusion process, we have implemented a voxel-level fusion approach instead of operating at the point level. To achieve this, the point clouds are first evenly divided into voxels, and the voxels are represented as $\{ V_i, i=1,2,3...E\}$, where $E$ is the overall count of the voxels. Specially, each voxel cell $V_{i}  = (\textbf{P}_i, \textbf{V}^{i}_{2D}, \textbf{V}^{i}_{3D}) \in \mathbb{R} ^{M \times(2m+3)}$ containing a fixed number of $M$ points, and each point attributes consist of $\textbf{P}_i$, $\textbf{V}^{i}_{2D}$, $\textbf{V}^{i}_{3D}$ which represent point coordinates, predicted 2D and 3D semantic segmentation results vector respectively. To ensure consistency across voxels, we utilize a sampling technique to maintain a fixed number of points in each voxel.
Subsequently, we utilize local and global feature aggregation techniques to calculate attention weights for each voxel, which help to determine the relative importance of 2D and 3D semantic segmentation labels. In other words, the AFF module chooses which category of 2D/3D semantic information should be trusted in the voxel.

 \begin{figure*}[h]
	\centering
  	\includegraphics[width=0.95\textwidth]{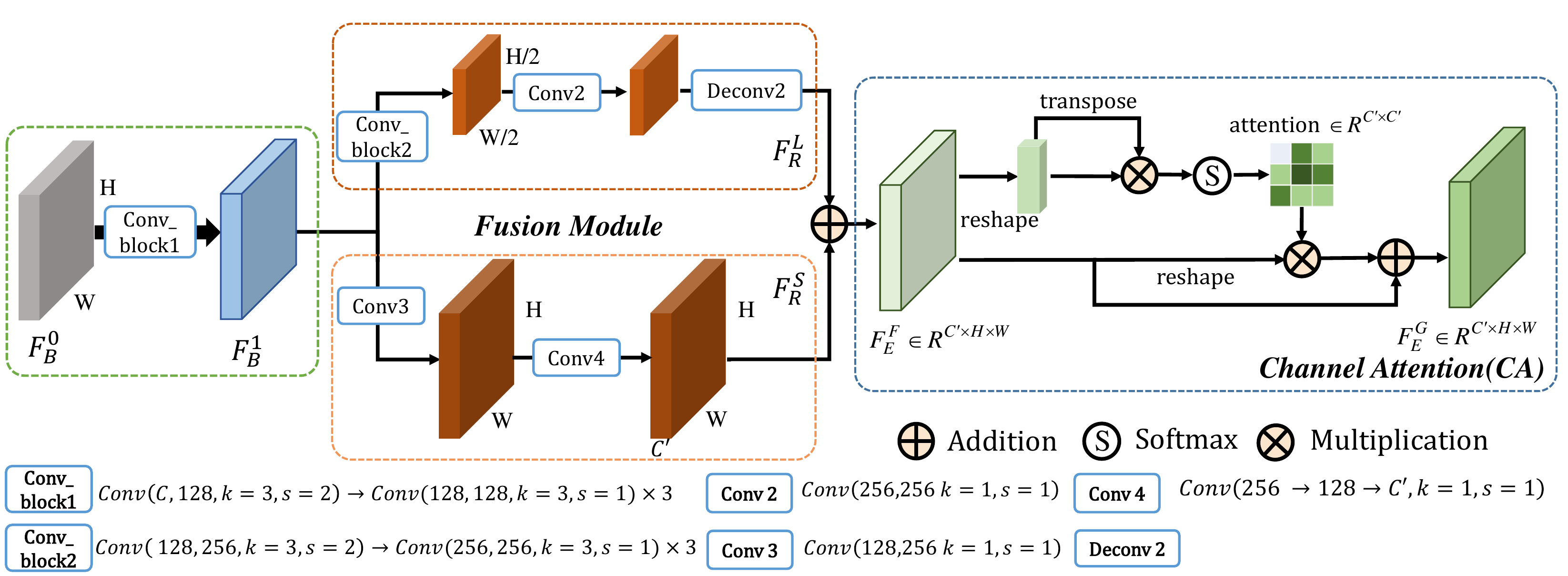}
	\centering
	\caption{An illustration of the proposed Deep Feature Fusion (DFF) module which includes one \textit{fusion} module and one \textit{channel attention} module respectively. The \textit{fusion} module includes two branches for producing features with different field-of-view.}
	\label{fig:deep_feature_fusion}
\end{figure*}

In order to get local features, we utilize a PointNet~\cite{qi2017pointnet}-like module to extract voxel-wise information within each non-empty voxel. Specifically, for the $i$-th voxel, the corresponding local feature is represented as: 
\begin{equation} \label{local_feature}
 V_i = f(p_1, p_2,  \cdots , p_M) =
 \max_{i=1,...,M} \{\text{MLP}_{l}(p_{i^{'}})\} \in \mathbb{R}^{C_1},
\end{equation}
{where $\{p_{i^{'}}, i^{'}=1,2,3...M\}$ indicates the LIDAR points insize each voxel}. $\text{MLP}_{l}(\cdot)$ and ${max}$ represent the muti-layer perception (MLP) and max-pooling module, respectively. Specifically, $\text{MLP}_{l}(\cdot)$ is composed of a linear layer, a batch normalization layer, an activation layer. Throughout the network, we got outputs for each voxel local feature with $C_1$ channels. For global feature information, we aggregate information based on the $E$ voxels where $E$ is the total number of voxels. First, we employ a $\text{MLP}_{g}(\cdot)$ module to rich each voxel features from $C_1$ dimensions to $C_2$. Then, we apply another PointNet-like module on all the voxels as follows expression:
\begin{equation} \label{global_feature}
V_{global} = f(V_1, V_2, \cdots ,V_E) = \max_{i=1,...,E} \{\text{MLP}_{g}(V_i)\} \in \mathbb{R}^{C_2}.
\end{equation}
To obtain the final fused local and global features, we expand the global feature vector $V_{global}$ to the same size as the number of voxels and then concatenate it with each local feature $V_i$. This operation creates a combined feature representation for each voxel that captures both local and global information as $V_{gl}\in \mathbb{R}^{E \times{(C_1 + C_2)}}$.

{After getting} fused features $V_{gl}$ from the network, we can estimate an attention score of two kinds of semantic information results for each point in voxel throughout another MLP module $\text{MLP}_{att}(\cdot)$ on $V_{gl}$ and a Sigmod activation function $\sigma(\cdot)$. Then, we multiply the resulting attention confidence score with corresponding one-hot semantic vectors for each voxel, as shown in Eq. (\ref{attention_2d}), Eq. (\ref{attention_3d}):
\begin{equation} \label{attention_2d}
	    \mathbf{V}^{i}_{a.S}  = \mathbf{V}^{i}_{2D} \times \sigma{(\text{MLP}_{att}(V^{i}_{gl})}) ,
\end{equation} 
\begin{equation}\label{attention_3d}
	    \mathbf{V}^{i}_{a.T} = \mathbf{V}^{i}_{3D} \times (1-\sigma{(\text{MLP}_{att}(V^{i}_{gl}})))
\end{equation} 
where $\mathbf{V}^{i}_{2D}$ and $\mathbf{V}^{i}_{3D}$ are the point labels in each voxel from 2D and 3D semantic segmentation results which are encoded with {one-hot} format. {The final semantic vector $V^{i}_{final}$ of each voxel can be obtained by concatenating or element-wise addition of $\mathbf{V}^{i}_{a.T}$ and $\mathbf{V}^{i}_{a.S}$.}

\begin{table*}[ht]
\centering
\renewcommand\arraystretch{1.1}
\resizebox{0.85\textwidth}{!}
{
\begin{tabular}{r|c|lll|lll|lll}
\hline
\multicolumn{1}{r|}{\multirow{2}{*}{$\textbf{Methods}$}} &\multicolumn{1}{c|}{\multirow{2}{*}{$\textbf{mAP}$ (Mod.)(\%)}} &
\multicolumn{3}{c|}{$\textbf{Car}$ $AP_{70}(\%)$} & 
\multicolumn{3}{c|}{$\textbf{Pedestrian}$ $AP_{70}(\%)$} & 
\multicolumn{3}{c}{$\textbf{Cyclist}$ $AP_{70}(\%)$}  \\ 

{} & {} & {Easy} & {Mod.} & {Hard} & 
{Easy} & {Mod.} & {Hard} & 
{Easy} & {Mod.} & {Hard}  \\ \hline  \hline
{SECOND\cite{yan2018second} } & {66.64} & 
{90.04} & {81.22} & {78.22} & 
{56.34} & {52.40} & {46.52} & 
83.94 & 66.31 & {62.37} \\
{SECOND $^{\ast}$} & {68.11} & 
{91.04} & {82.31} & {79.31} & 
 {59.28} & {54.18} & {50.20} & 
85.11 & 67.52 & {63.36} \\ 
{Improvement} & \R{+1.47} & 
\R{+1.00} & \R{+1.09} & \R{+1.09} & 
\R{+2.94} & \R{+1.78} & \R{+3.68} & 
\R{+1.17} & \R{+1.54} & \R{+0.99} \\ \hline 

{Pointpillars\cite{lang2019pointpillars} } & {62.90} 
& {87.59} & {78.17} & {75.23} & 
 {53.58} & {47.58} & {44.04} & 
{82.21} & {62.95} & {58.66} \\
{PointPillars$^{\ast}$} & {65.78} &
{89.58} & {78.60} & {75.63} & 
{60.22} & {54.23} & {49.49} & 
84.83 & 64.50 & {60.17} \\ 

{Improvement} & \R{+2.88} & 
\R{+1.99} & \R{+0.43} & \R{+0.4} & 
\R{+6.64} & \R{+6.65} & \R{+5.45} & 
\R{+2.62} & \R{+1.55} & \R{+1.51} \\ \hline

{PV-RCNN\cite{shi2020pv}} & {71.82} &
{92.23} & {83.10} & {82.42} & 
{65.68} & {59.29} & {53.99} & 
91.57 & 73.06 & {69.80} \\
{PV-RCNN$^{\ast}$} & {73.95} & 
{91.85} & {84.59} & {82.66} & 
{69.12} & {61.61} & {55.96} & 
94.90 & 75.65 & {71.03} \\ 
{Improvement} & \R{+2.13} & 
\G{-0.38} & \R{+1.49} & \R{+0.24} & 
\R{+3.44} & \R{+2.32} & \R{+1.97} & 
\R{+3.33} & \R{+2.59} & \R{+1.23} \\ \hline
\end{tabular}
}
\caption{3D object detection evaluation on KITTI ``val'' split on different baseline approaches, where $^{\ast}$ represents the boosted baseline by adding the proposed fusion modules. ``Easy'', ``Mod.'' and ``Hard'' represent the three difficult levels defined by official benchmark and $\textbf{mAP}$ (Mod.) represents the average $\textbf{AP}$ of ``Car'', ``Pedestrian'' and ``Cyclist'' on ``Mod.'' level. For easy understanding, we also highlight the improvements with different colors, where red represents an increase and green represents a decrease compared to the baseline method. This table is better to be viewed in color mode.}
\label{tab:kitti_3D_detection}
\end{table*}

\begin{table*}[!ht]
\centering
\renewcommand\arraystretch{1.1}
\resizebox{0.85\textwidth}{!}
{
    \begin{tabular}{r|c| ccc | ccc| ccc}
    \hline
    \multirow{2}{*}{$\textbf{Methods}$} &\multirow{2}{*}{$\textbf{mAP}$ (Mod.)(\%)} &
    \multicolumn{3}{c|}{$\textbf{Car}$ $AP_{70}(\%)$} &
    \multicolumn{3}{c|}{$\textbf{Pedestrian}$ $AP_{70}(\%)$} & 
    \multicolumn{3}{c}{$\textbf{Cyclist}$ $AP_{70}(\%)$}  \\ 
    {} & {} & {Easy} & {Mod.} & {Hard} & 
    {Easy} & {Mod.} & {Hard} & 
    {Easy} & {Mod.} & {Hard}  \\ \hline  \hline 
    {SECOND\cite{yan2018second} }  & {71.95}  
    & {92.31} & {88.99} & {86.59}  
    & {60.5} & {56.21} & {51.25}  
    & {87.30} & {70.65} & {66.63} \\
    SECOND $^{\ast}$ & {73.72}  
    & {94.70} & {91.62} & {88.35}  
    & {63.95} & {59.66} & {55.81}  
    & {91.95} & {73.28} & {67.75} \\
    {Improvement} & \RED{{+2..90}} 
    & \RED{+2.39} & \RED{+2.63}  & \RED{+1.76}  
    & \RED{+3.45} & \RED{+3.45} & \RED{+4.56} 
    & \RED{+4.65} & \RED{+2.63} & \RED{+1.12} \\ \hline
    {Pointpillars\cite{lang2019pointpillars} } & {69.18}  
    & {92.50} & {87.80} & {87.55}  
    & {58.58} & {52.88} & {48.30} 
    & {86.77} & {66.87} & {62.46} \\
    PointPillars $^{\ast}$  & {72.13}  
    & {94.39} & {87.65} & {89.86}  
    & {64.84} & {59.57} & {55.16}  
    & {89.55} & {69.18} & {64.65} \\
    {Improvement} & \RED{{+2.95}}  
    & \RED{+1.89} & \GRE{-0.15} & \RED{+2.31}  
    & \RED{+6.26} & \RED{+6.69} & \RED{+6.86}  
    & \RED{+2.78} & \RED{+2.31} & \RED{+2.19} \\ \hline
    {PV-RCNN\cite{shi2020pv}}  & {76.23}  
    & {94.50} & {90.62} & {88.53}  
    & {68.67} & {62.49} & {58.01} 
    & {92.76} & {75.59} & {71.06} \\
    PV-RCNN $^{\ast}$ & {78.17}  
    & {94.86} & {90.87} & {88.88}  
    & {71.99} & {64.71} & {59.01} 
    & {96.35} & {78.93} & {74.51} \\
    {Improvement} & \RED{{+1.94}}  
    & \RED{+0.36} & \RED{+0.25} & \RED{+0.35}
    & \RED{+3.32} & \RED{+2.22} & \RED{+1.00} 
    & \RED{+3.59} & \RED{+3.34} & \RED{+3.45} \\ \hline
    \end{tabular}
}
\caption{Evaluation of bird's-eye view object detection on KITTI ``val'' split with different baselines. Similar to the 3D detection, the red color represents an increase and the green color represents a decrease compared to the baseline method. This table is also better to be viewed in color mode.}
\label{tab:KITTI_BEV_detection}
\end{table*}

\subsection{Deep Feature Fusion Module} \label{subsec:deep_fusion}
In AD scenarios, determining the identity and location of objects is crucial for the subsequent planning and control modules. Therefore, it is not only necessary to recognize what objects are present, but also where they are located. In typical object detection frameworks, they correspond to the classification and regression branches respectively. Empirically, global context information is important to recognize the specific class attributes. On the contrary, the object's attributes (e.g., dimension, orientation, and precise location, etc) regression branch prioritize the capture of detailed spatial information around the ROI (region of interest) in a relatively small range. For accurate kinds of size object detection, therefore various scales receptive fields are necessary. This issue has been considered in most object detection frameworks. However, how to use various fields of view fields in an efficient way is vitally important.

To handle this issue, a specific DFF module is proposed to aggregate both the low-level and high-level features with different receptive fields. 
The \figref{fig:deep_feature_fusion} shows the architecture of the DFF module. First, the backbone features $F^{0}_B$ from the feature extractor pass the \textit{Conv\_block1} with several convolution layers to obtain the $F^{1}_B$ as a basic feature. 
Here, \textit{Conv\_block1} has four Conv modules and the first $conv$ module takes $C$ channels as input and outputs 128 channels, and the following three $convs$ share the same input {channels} and output channels. For each $conv$ module in the \figref{fig:deep_feature_fusion}, it consists one $Con2d$, a batch normalization layer, and a Rectified Linear Unit (ReLU) activation layer. For easy understanding, we have given the stride and kernel size for each $conv$ operation at the bottom of \figref{fig:deep_feature_fusion}. Then, the feature $F^{1}_B$ will pass two branches to obtain the features with different receptive fields. For one branch, the features are down-sampled into $1/2$ size with $Conv-block2$ first and then pass the $Conv2$ operation. Finally, the outputs are up-sampled into the feature map $F_{R}^{L}\in \mathbb{R}^{H \times W \times C' }$ with $Deconv2$. For the other branch, $F^{1}_{B}$ will pass $Covn3$ and $Conv4$ to obtain the features $F_{R}^{S}$ consecutively. And the shape of the output $F_{R}^{L}$ is the same as $F_{R}^{S}$. Furthermore, we use the addition operation for fusing different level perception field features to improve the feature representation. 

After adding the high-level and low-level features element-wisely, a channel-attention (CA) module like \cite{fu2019dual} is employed to further fuse both of them. The architecture of the module can be found in \figref{fig:deep_feature_fusion}, which is named \textbf{CA}. Usually, the channel feature from low-level to high-level throughout the backbone will take the loss of information. In order to minimize this influence, we utilize the CA module to selectively emphasize interdependent channel maps by integrating relative features among all channel maps. Specially, the enhanced feature  $F_{E}^{F} \in \mathbb{R}^{C \times H \times W}$ is reshaped to $F_{E}^{F} \in \mathbb{R}^{C \times{N}}$, where $N$ is the pixel numbers of each channel.
The channel dependencies for feature $F^{F}_E$ are captured using the similar self-attention mechanism, followed by updating the enhanced feature through a weighted sum of all channel maps. The process as shown in Eq.(\ref{CA1}), Eq.(\ref{CA2}):
\begin{equation} \label{CA1}
x_{j i}=\frac{\exp \left({F^{F}_E}_{i} \cdot {F^{F}_E}_{j}\right)}{\sum_{i=1}^{C} \exp \left({F^{F}_E}_{i} \cdot {F^{F}_E}_{j}\right)}
\end{equation} 

\begin{equation} \label{CA2}
{F^{G}_E}_{j}=\beta \sum_{i=1}^{C}\left(x_{j i} {{F^{F}_E}}_{i}\right)+{{F^{F}_E}}_{j}
\end{equation} 
where $x_{j i}$ measures the $i^{th}$ channel's impact on the $j^{th} $ channel. In addition, we perform a matrix multiplication between the transpose of $x$, which channel attention map $\mathbf{x} \in \mathbb{R}^{C \times C}$. Moreover, $\beta $ is a scale parameter, which gradually learns a weight from 0. The Eq. (\ref{CA2}) shows that the final feature of each channel is a weighted sum of the features of all channels and original features, which models the long-range semantic dependencies between feature maps. Finally, $F^{G}_E$ is taken as the inputs for the following classification and regression heads.

\subsection{3D Object Detection Framework} \label{subsec:3d_detector}

{The proposed AAF and DFF modules are }detector independent and any off-the-shelf 3D object detectors can be directly employed as the baseline of our proposed framework. The 3D detector receives the points or voxels produced by the AAF module as inputs and can keep backbone structures unchanged to {obtain the backbone features. Then, the backbone features are boosted by passing the proposed DFF module. Finally, detection results are generated from the classification and regression heads.}


\section{Experimental Results} \label{sec:experiments}

To verify the effectiveness of our proposed framework, we evaluate it on two large-scale 3D object detection dataset in AD scenarios as KITTI \cite{geiger2012we} and nuScenes \cite{caesar2020nuscenes}. Furthermore, {the proposed modules is also evaluated on different kinds of baselines} for verifying its generalizability, including SECOND \cite{yan2018second}, PointPilars \cite{lang2019pointpillars} and PVRCNN \cite{shi2020pv}, etc.

\subsection{Evaluation on KITTI Dataset} \label{subsec:eval_kitti}
 
\textit{KITTI} is one of the most popular benchmarks for 3D object detection in AD, which contains 7481 samples for training and 7518 samples for testing. The objects in each class are divided into three difficulty levels as ``easy'', ``moderate'', and ``hard'', according to the object size, the occlusion ratio, and the truncation level. Since the ground truth annotations of the test samples are not available and the access to the test server is limited, we follow the idea in~\cite{chen2017multi} and split the training data into ``train'' and ``val'' where each set contains 3712 and 3769 samples respectively. In this dataset, both the LiDAR point clouds and the RGB images have been provided. In addition, both the intrinsic parameters and extrinsic parameters between different sensors have been given. 

\begin{table*}[ht!]
\centering
\normalsize
\resizebox{1\textwidth}{!}
{
	\begin{tabular}{l | c | c | c c c c c c c c c c}
	\hline
    {\multirow{2}{*}{Methods}} &
    {\multirow{2}{*}{\textbf{NDS} (\%)}} &
    {\multirow{2}{*}{\textbf{mAP} (\%)}} & 
    \multicolumn{10}{c}{\textbf{AP} (\%)} \\
    {} &  {} & {} & {Car} & {Pedestrian} &{Bus} &{Barrier} & {T.C.} &
    {Truck} & {Trailer} & {Moto.} & {Cons.} &
    {Bicycle} \\ \hline \hline
	SECOND \cite{yan2018second}  &61.96 &50.85 &81.61  &77.37 & 68.53 &57.75  &56.86  &51.93 &38.19 & 40.14 &17.95 & 18.16 \\
	SECOND$^{\ast}$ & 67.61  & 62.61  & 84.77  & 83.36 &72.41 &63.67 &74.99 &60.32 &42.89 &66.50 &23.59  &54.62  \\
	Improvement $\uparrow$ & \textcolor{red}{\textbf{+5.65}} & \textcolor{red}{\textbf{+11.76}} & \textcolor{red}{+3.16} & \textcolor{red}{+5.99} & \textcolor{red}{+3.88} & \textcolor{red}{+5.92} & \textcolor{red}{+18.13} & \textcolor{red}{+8.39} & \textcolor{red}{+4.70} & \textcolor{red}{+26.36} & \textcolor{red}{+5.64}& \textcolor{red}{+36.46} \\
    \hline
	PointPillars \cite{lang2019pointpillars}  & 57.50 &43.46 &80.67 &70.80 &62.01 &49.23 &44.00 &49.35 &34.86 &26.74 &11.64 &5.27 \\
	PointPillars$^{\ast}$ &66.43 & 61.75 &84.79 &83.41 &70.52 &59.42 &75.43 &57.50 &42.32 &66.97 &22.43 &54.68   \\
	Improvement $\uparrow$ & \textcolor{red} {\textbf{+8.93}} & \textcolor{red}{\textbf{+18.29}} & 
	\textcolor{red}{+4.12} & \textcolor{red}{+12.61} & \textcolor{red}{+8.51 } & \textcolor{red}{ +10.19} & \textcolor{red}{+31.43} & \textcolor{red}{+8.15} & \textcolor{red}{+7.46} & \textcolor{red}{+40.23}  & \textcolor{red}{+10.79} & \textcolor{red}{+49.41} \\
    \hline
	CenterPoint \cite{yin2021center} & 64.82 &56.53 &84.73 &83.42 &66.95 &64.56 &64.76 &54.52 &36.13 &56.81 &15.81 &37.57\\
	CenterPoint$^{\ast}$ &69.91 &65.85  &87.00 &87.95 & 71.53 &67.14 &79.89 &61.91 & 41.22 & 73.85 &23.97 &64.06\\
	Improvement $\uparrow$ & \textcolor{red}{\textbf{+5.09}} & \textcolor{red}{\textbf{+6.81}} & \textcolor{red}{+2.77} & \textcolor{red}{+4.53} & \textcolor{red}{+4.58} & \textcolor{red}{+2.58} & \textcolor{red}{+15.13} & \textcolor{red}{+7.39} & \textcolor{red}{+5.09} & \textcolor{red}{+17.04} &  \textcolor{red}{+8.16}&\textcolor{red}{+26.49} \\
    \hline
	\end{tabular}
}
\caption{\normalfont Evaluation results on nuScenes validation dataset. ``NDS'' and ``mAP'' mean nuScenes detection score and mean Average Precision. ``T.C.'', ``Moto.'' and ``Cons.'' are short for ``traffic cone'', ``motorcycle'', and ``construction vehicle'' respectively. `` * '' denotes the improved baseline by adding the proposed fusion module. The red color represents an increase compared to the baseline. This table is also better to be viewed in color mode.}
\label{tab:eval_on_nuscenes_val}
\end{table*}

\noindent\textbf{Evaluation Metrics.} {We follow the official metrics provided by the KITTI for evaluation. ${AP}_{70}$ is used for ``Car'' category while ${AP}_{50}$ is used for ``Pedestrain'' and ``Cyclist''. }
Specifically, before the publication of \cite{simonelli2019disentangling}, the KITTI official benchmark used the 11 recall positions for comparison. After that, the official benchmark changes the evaluation criterion from 11-points to 40-points because the latter one is proved to be more stable than the former \cite{simonelli2019disentangling}. Therefore, we use the 40-points criterion for all the experiments here. In addition, similar to \cite{vora2020pointpainting}, the average AP (mAP) of three Classes for ``Moderate'' is also taken as an indicator for evaluating the average performance on all three classes. 

\noindent\textbf{Baselines.} Three different baselines have been used for evaluation on KITTI: 
\begin{enumerate}
    \item \textit{SECOND \cite{yan2018second}} is the first to employ the sparse convolution on the voxel-based 3D object detection framework to accelerate the efficiency of LiDAR-based 3D object detection.  
    \item \textit{PointPillars \cite{lang2019pointpillars}} {is proposed to further improve the detection efficiency by dividing the point cloud into vertical pillars rather than voxels. For each pillar, \textit{Pillar Feature Net} is applied to extract the point-level feature.} 
    \item\textit{PV-RCNN \cite{shi2020pv}} is a hybrid point-voxel-based 3D object detector, which can utilize the advantages from both the point and voxel representations.
\end{enumerate}

\noindent\textbf{Implementation Details.} DeeplabV3+ \cite{deeplabv3plus2018} and Cylinder3D \cite{zhu2021cylindrical} are employed for 2D and 3D scene parsing respectively. More details, the DeeplabV3+ is pre-trained on Cityscape \footnote{\url{https://www.cityscapes-dataset.com}}, and the Cylinder3D is trained on KITTI point clouds by taking points in 3D ground trues bounding box as foreground annotation. For AAF module, $m = 4$, $C_1 = 64,C_2 = 128$, respectively. {The voxel size for PointPillars and SECOND are $0.16m \times 0.16m \times 4m$ and $0.05m \times 0.05m \times 0.1m$} respectively. In addition, both of the two baselines use the same optimization (e.g., AdamW) and learning strategies (e.g., one cycle) for all the experiments.
{In the DFF module, we set $C = 256$ and $C' = 512$ for \textit{SECOND} and \textit{PV-RCNN} framework and $C = 64$ and $C' = 384$ for \textit{PointPillars} network. The kernel and stride size are represented with $k$ and $s$ in Fig.\ref{fig:deep_feature_fusion}, respectively.}

{The proposed approach is implemented with PaddlePaddle \cite{PaddlePaddle_2019} and all the methods are trained on NVIDIA Tesla V100 with 8 GPUs. The AdamW is taken as the optimizer and the one-cycle learning strategy is adopted for training the network. For \textit{SECOND} and \textit{PointPillars}, the batch size is 4 per GPU and the maximum learning rate is 0.003 while the bath size is 2 per GPU and the maximum learning rate is 0.001 for \textit{PV-RCNN}.}

\noindent\textbf{Quantitative Evaluation.} We illustrate the results on 3D detection and Bird's-eye view in \tabref{tab:kitti_3D_detection} and \tabref{tab:KITTI_BEV_detection} respectively. 
{From the table, we can clearly see that remarkable improvements have been achieved on the three baselines across all the categories.}
Taking Pointpillars as an example, the proposed method has achieved 0.43\%, 6.65\%, 1.55\% points improvements on ``Car'', ``Pedestrian'' and ``Cyclist'' respectively. Interestingly, compared to ``Car'', ``Pedestrian'' and ``Cyclist'' give much more improvements by the fusion painting modules. 

\begin{table*}[ht!]
    \centering
    \Large
\resizebox{0.9\textwidth}{!}{

\begin{tabular}{ l| c| c| c| c c c c c c c c c c}
    \hline
    {\multirow{2}{*}{Methods}}  & 
    {\multirow{2}{*}{Modality}} & {\multirow{2}{*}{\textbf{NDS}(\%)}} & 
    {\multirow{2}{*}{\textbf{mAP}(\%)}} & 
    \multicolumn{10}{c}{\textbf{AP} (\%)} \\ 
    {} & {} & {} & {} & {Car} & {Truck} & {Bus} & {Trailer} & {Cons} & {Ped} & {Moto} & {Bicycle} & {T.C} & {Barrier} \\ \hline \hline
      PointPillars \cite{vora2020pointpainting} & L & 55.0 & 40.1 & 76.0 & 31.0 & 32.1 & 36.6 & 11.3 & 64.0 & 34.2 & 14.0 & 45.6 & 56.4 \\
    VoxelNeXt \cite{VoxelNeXt} & L &  70.0 & 64.5& 84.6 &53.0& 64.7& 55.8& 28.7& 85.8& 73.2& 45.7& 79.0& \textbf{74.6} \\
    Focals Conv \cite{focalsconv} &  L  &  70.0 & 63.8& 86.7& 56.3& \textbf{67.7} &59.5& 23.8& 87.5& 64.5& 36.3& 81.4 &74.1 \\
    3DSSD \cite{yang20203dssd} & L  & 56.4 & 46.2 & 81.2  & 47.2  & 61.4  &  30.5 & 12.6 &  70.2 & 36.0  &  8.6 &  31.1 &  47.9 \\
    CBGS \cite{zhu2019class} & L & 63.3 & 52.8 & 81.1 & 48.5 & 54.9 & 42.9 & 10.5 & 80.1 & 51.5 & 22.3 & 70.9 & 65.7 \\
    HotSpotNet \cite{chen2020object} & L & 66.0 & 59.3 & 83.1  & 50.9 & 56.4  & 53.3  &  {23.0} &  81.3 & {63.5}  &  36.6 &  73.0 &  71.6 \\
    VP-Net\cite{vpnet} & L & 67.5 & 57.5 & 77.9 & 52.9 & 63.8 & 56.9 & 24.3 & 79.4 & 46.8 & 32.6 & 70.7 & 66.2 \\
    CVCNET\cite{chen2020every} & L & 66.6 & 58.2 & 82.6 & 49.5 & 59.4 & 51.1 & 16.2 & {83.0} & 61.8 & {38.8} & 69.7 & 69.7 \\
    CenterPoint \cite{yin2021center} & L & {67.3} & {60.3} & {85.2} & {53.5} & {63.6} & {56.0} & 20.0 & 54.6 & 59.5 & 30.7 & {78.4} & 71.1 \\
    PointPainting \cite{vora2020pointpainting} & L \& C & 58.1 & 46.4 & 77.9 & 35.8 & 36.2 & 37.3 & 15.8 & 73.3 & 41.5 & 24.1 & 62.4 & 60.2 \\
    UVTR \cite{uvtr} &  L \& C &  71.1 &67.1& \textbf{87.5} &56.0& 67.5& 59.5&\textbf{33.8}& 86.3 &73.4& \textbf{54.8}& 79.6 &73.0 \\
    3DCVF \cite{DBLP:conf/eccv/YooKKC20} & L \& C & 62.3 & 52.7 & 83.0 & 45.0 & 48.8 & 49.6 & 15.9 & 74.2 & 51.2 & 30.4 & 62.9 & 65.9 \\ \hline
    {Our proposed} & L \& C & \textbf{71.6} & \textbf{68.2} & 87.1 & \textbf{60.8} & 66.5 & \textbf{61.7} & 30.0 & \textbf{88.3} & \textbf{74.7} & 53.5 & \textbf{85.0} & 71.8  \\  
    \hline
    \end{tabular}

}
\caption{Comparison with other SOTA methods on the nuScenes 3D object detection testing benchmark. ``L'' and ``C'' in the modality column represent LiDAR and Camera sensors respectively. For easy understanding, the highest score in each column is shown in bold font. To be clear, only the results with publications are listed here.}
\label{tab:test_tabel}
\end{table*}

\subsection{Evaluation on the nuScenes Dataset} \label{subsec:nuScenes}
The nuScenes \cite{caesar2020nuscenes} is a recently released large-scale (with a total of 1,000 scenes) AD benchmark with different kinds of information including LiDAR point cloud, radar point, Camera images, and High Definition Maps, etc. For a fair comparison, the dataset has been divided into ``train'', ``val'', and ``test'' three subsets officially, which includes 700 scenes (28130 samples), 150 scenes (6019 samples), and 150 scenes (6008 samples) respectively. Objects are annotated in the LiDAR coordinate and projected into different sensors' coordinates with pre-calibrated intrinsic and extrinsic parameters. For the point clouds stream, only the keyframes (2fps) are annotated. With a 32 lines LiDAR scanner, each frame contains about {300,000} points for 360-degree viewpoint. For the object detection task, the obstacles have been categorized into 10 classes as ``car'', ``truck'', ``bicycle'' and ``pedestrian'' etc. Besides the point clouds, the corresponding RGB images are also provided for each keyframe, and for each keyframe, there are 6 cameras that can cover 360 fields of view.

\noindent\textbf{Evaluation Metrics.} The evaluation metric for nuScenes is totally different from KITTI and they propose to use mean Average Precision (mAP) and nuScenes detection score (NDS) as the main metrics. Different from the original mAP defined in \cite{everingham2010pascal}, nuScenes consider the BEV center distance with thresholds of \{0.5, 1, 2, 4\} meters, instead of the IoUs of bounding boxes. NDS is a weighted sum of mAP and other metric scores, such as average translation error (ATE) and average scale error (ASE). For more details about the evaluation metric please refer to \cite{caesar2020nuscenes}.

\noindent\textbf{Baselines.} We have integrated the proposed module on three different SOTA baselines to verify its effectiveness. Similar to the KITTI dataset, both the \textit{SECOND} and \textit{PointPillars} are employed and the other detector is \textit{CenterPoint} \cite{yin2021center}. which is the first anchor-free-based 3D object detector for LiDAR-based 3D object detection.  


\noindent\textbf{Implementation Details} HTCNet \cite{chen2019hybrid} and Cylinder3D \cite{zhu2021cylindrical} are employed here for obtaining the 2D and 3D semantic segmentation results respectively. We used the HTCNet model trained on nuImages \footnote{\url{https://www.nuscenes.org/nuimages}} dataset directly for generating the semantic labels. For Cylinder3D, we train it directly on the nuScenes 3D object detection dataset while the point cloud semantic label is produced by taking the points inside each bounding box. In AAF module, $m = 11$, $C_1 = 64$ and  $C_2 = 128$ respectively. 
{In DFF module, $C = 256 $ and $C' = 512$ for \textit{SECOND} and \textit{CenterPoint} while $C = 64$ and $C' = 384$ for \textit{PointPillars}. The setting for kernel size $k$ and stride $s$ are given in Fig. \ref{fig:deep_feature_fusion} and same for all three detectors.} The voxel size for PointPillars, SECOND and CenterPoint are $0.2m \times 0.2m \times 8 m$, $0.1m \times 0.1m \times0.2m$ and $0.075m \times 0.075m \times 0.075m$, respectively. We use AdamW \cite{ loshchilov2019decoupled} as the optimizer with the max learning rate is 0.001. Following \cite{caesar2020nuscenes}, 10 previous LiDAR sweeps are stacked into the keyframe to make the point clouds more dense.



{All the baselines are trained on NVIDIA Tesla V100 (8 GPUs) with a batch size of 4 per GPU for 20 epochs. AdamW is taken as the optimizer and the one-cycle learning strategy is adopted for training the network with the maximum learning rate is 0.001.} 



\noindent\textbf{Quantitative Evaluation.}  
{The proposed framework has been evaluated on nuScenes benchmark for both ``val'' and ``test'' splits.} {The results of comparison with three baselines are given in Tab. \ref{tab:eval_on_nuscenes_val}. From this table, we can see that significant improvements have been achieved on both the $\text{mAP}$ and $\text{NDS}$ across all the three baselines. For the \textit{PointPillars}, the proposed module gives \textbf{8.93} and \textbf{18.29} points improvements on $\text{NDS}$ and $\text{mAP}$ respectively. For \textit{SECOND}, the two values are \textbf{5.65} and \textbf{11.76} respectively. Even for the strong baseline \textit{CenterPoint}, the proposed module can also give \textbf{5.09} and \textbf{9.32} points improvements.}
In addition, we also find that the {categories which share small sizes} such as ``Traffic Cone'', ``Moto'' and ``Bicycle'' have received more improvements compared to other categories. Taking ``Bicycle'' as an example, the $\text{mAP}$ has been improved by {36.46\%}, {49.41\%} and {26.49\%} compared to \textit{SECOND}, \textit{PointPillar} and \textit{Centerpoint} respectively. {This phenomenon can be explained as these categories with small sizes are hard to be recognized in point clouds because of a few LiDAR points on them. In this case, the semantic information from the 2D/3D parsing results is extremely helpful for improving 3D object detection performance. It should be emphasized that the result in Tab. \ref{tab:eval_on_nuscenes_val} without any Test Time Augmentation(TTA) strategies.}


To compare the proposed framework with other SOTA methods, we submit our results (adding fusion modules on \textit{CenterPoint}) on the nuScenes evaluation sever\footnote{\url{https://www.nuscenes.org/object-detection/}} for test split. The detailed results are given in Tab. \ref{tab:test_tabel}. From this table, we can find that the proposed method achieves the best performance on both the $\text{mAP}$ and $\text{NDS}$ scores. Compared to the baseline \textit{CenterPoint}, 4.3 and 7.9 points improvements have been achieved by adding our proposed fusion modules. For easy understanding, we have highlighted the best performances in bold in each column. It should be noted that the results in Tab. \ref{tab:test_tabel} the Test Time Augmentation(TTA) strategy including multiple flip and rotation operations is applied during the inference time which is 1.69 points higher than the origin method of ours.

\subsection{Ablation Studies} \label{subsec:ablation_study}


{To verify the effectiveness of different modules, a series of ablation experiments have been designed to analyze the contribution of each component for the final detection results. Specially, three types of experiments are given as \textit{different semantic representations} and \textit{different fusion modules} and \textit{effectiveness of channel attention}.
}

\begin{table}[t]
\centering
\small
\renewcommand\arraystretch{1.1}
\resizebox{0.4\textwidth}{!}
{   
    \begin{tabular}{r | cccc}
    \hline
    $\textbf{Representations}$ & $\textbf{mAP }$ (\%) & $\textbf{NDS}$ (\%)  \\ \hline 
    {PointPillar \cite{lang2019pointpillars}} & 43.46 & 57.50  \\ 
    {Semantic ID}    & 50.96 (+7.50)  & 60.69 (+3.19)   \\ 
    {Onehot Vector}   & 52.18 (+8.72)  & 61.59 (+4.09)   \\ 
    {Semantic Score}  & 53.10 (+9.64)  & 62.20 (+4.70)   \\ \hline
    \end{tabular}
}
\caption{Ablation studies on the nuScenes \cite{geiger2012we} dataset for fusing the semantic results with different representations.}
\label{tab:ablation_dif_semantic_reps}
\end{table}

\noindent\textbf{Different Semantic Representations.} First of all, we investigate the influence of the different semantic result representations on the final detection performance. Three different representations as ``Semantic ID'', ``Onehot Vector'' and ``Semantic Score'' are considered here. For ``Semantic ID'', the digital class ID is used directly and for the ``Semantic Score'', the predicted probability after the Softmax operation is used. In addition, to convert the semantic scores to a ``Onehot Vector'', we assign the class with the highest score as ``1'' and keep other classes as ``0''. Here, we just add the semantic feature to the original $(x, y, z)$ coordinates with concatenation operation and \textit{PointPillars} is taken as the baseline on nuScenes dataset due to its efficiency of model iteration. From the results in \tabref{tab:ablation_dif_semantic_reps}, we can easily find that the 3D semantic information can significantly boost the final object detection performance regardless of different representations. More specifically, the ``Semantic Score'' achieves the best performance among the three which gives 9.64 and 4.70 points improvements for $\textbf{mAP}$ and $\textbf{NDS}$ respectively. We guess that the semantic score can provide more information than the other two representations because it not only provides the class ID but also the confidence for all the classes.    

\noindent\textbf{Different Fusion Modules.} We also execute different experiments to analyze the performances of each fusion module on both KITTI and nuScene  dataset. The \textit{SECOND} is chosen as the baseline for KITTI while all the three detectors are verified on nuScene. 
The results are given in \tabref{tab:ablation_kitti_table} and \tabref{tab:ablation_nuScene_table} for KITTI and nuScenes respectively. To be clear, the ``3D Sem'' and ``2D Sem.'' represent the 2D and 3D parsing results. ``AAF'' represents the fusion of the 2D and 3D semantic information with the proposed adaptive attention module and the ``AAF + DFF'' represents the module with both the two fusion strategies.

\begin{table}[t]
\centering
\small
\resizebox{0.5\textwidth}{!}
{%
    \begin{tabular}{r | ccc|l}
    \hline
\multirow{2}{*}{Strategies} & \multicolumn{3}{c|}{AP(\%)}  & \multirow{2}{*}{mAP(\%)} \\
                            & \multicolumn{1}{l}{Car(Mod.)} & \multicolumn{1}{l}{Ped.(Mod.)}   & \multicolumn{1}{l|}{Cyc.(Mod.)}  &  \\ \hline
    {SECOND \cite{yan2018second}} & 88.99 & 56.21 & 70.65 & 71.95 \\ 
    {3D Sem.}             & + 0.81  &  + 0.70  &  + 0.63  & + 0.71  \\ 
    {2D Sem.}    &  + 1.09 &  + 1.35  &  + 1.20  &  + 1.41  \\ 
    {AAF}  & + 1.62   &  + 1.78  &  + 1.91  & + 1.77  \\ 
    {AAF \& DFF} & + 2.63   &  + 3.45  &  + 2.63  &  + 2.90 \\ \hline
    \end{tabular}
}
\caption{Evaluation of ablation studies on the public KITTI \cite{geiger2012we} dataset. Similar to PointPainting \cite{vora2020pointpainting}, we provide the 2D BEV detection here. To be clear, only the results of ``Mod'' have been given and \textbf{mAP} is the average mean of all the three categories.}
\label{tab:ablation_kitti_table}
\end{table}

\begin{table}[ht!]
\centering
\setlength{\tabcolsep}{2pt}
\renewcommand\arraystretch{1.1}
\resizebox{0.5\textwidth}{!}
{%
    \begin{tabular}{r | cc | cc |cc}
    \hline
    \multicolumn{1}{r|}{\multirow{2}{*}{\textbf{Strategies}}} & 
    \multicolumn{2}{c|}{\textbf{SECOND}}   & 
    \multicolumn{2}{c|}{\textbf{PointPillars}} & 
    \multicolumn{2}{c}{\textbf{CenterPoint}}  \\ 
   {}  & \textbf{mAP}(\%) & \textbf{NDS}(\%) &  \textbf{mAP}(\%) & \textbf{NDS}(\%)& \textbf{mAP}(\%)& \textbf{NDS}(\%) 
   \\ \hline  
    {SECOND}         &  50.85 &  61.96 &   43.46 &  57.50    &  56.53 & 64.82\\ 
    {3D Sem.}        &   + 3.60&   + 1.45  & + 8.72  &  + 4.09 & + 4.59 & + 1.86 \\ 
   {2D Sem.}         &   + 8.55&   + 4.07  &  +15.64   &+ 7.55    & + 6.28 & + 3.67 \\ 
    {AAF}   & + 11.30  &  + 5.45   &  +17.31   & + 8.54   & + 8.46& + 4.56 \\ 
   {AAF \& DFF}  &  + 11.76 &  + 5.65   &  +  18.19  & + 8.93     & + 9.32 & + 5.09\\ \hline
    \end{tabular}
}
\caption{Ablation studies for different fusion strategies on the nuScenes benchmark. The first row is the performance of the baseline method and the following values are the gains obtained by adding different modules.}
\label{tab:ablation_nuScene_table}
\end{table}

In \tabref{tab:ablation_kitti_table}, the first row is the performance of the baseline method and the following values are the gains obtained by adding different modules. From the table, we can obviously find that all the modules give positive effects on all three categories for the final detection results. For all the categories, the proposed modules give 2.9 points improvements averagely and ``Pedestrian'' achieves the most gain which achieves 3.45 points. Furthermore, we find that deep fusion gives the most improvements compared to the other three modules in this dataset.   

\tabref{tab:ablation_nuScene_table} gives the results on the nuScenes dataset. 
{To be clear, a boosted version of baseline is presented here than in \cite{xu2021fusionpainting} by fixing the sync-bn issue in the source code.} 
From this table, we can see that both the 2D and 3D {semantic information} can significantly boost the performances while the 2D information provide more improvements than the 3D information. This phenomenon can be explained as that the 2D texture information can highly benefit the categories classification results which is very important for the final \textbf{mAP} computation. In addition, by giving the 2D information, the recall ability can be improved especially for objects in long distance with a few scanned points. In other words, the advantage for 3D semantic information is that the point clouds can well handle the occlusion among objects which are very common in AD scenarios which is hard to deal with in 2D. After fusion, all the detectors achieve much better performances than only one semantic information (2D or 3D).

Furthermore, a deep fusion module is also proposed to aggregate the backbone features to further improve the performance. From the Tab. \tabref{tab:ablation_kitti_table} and \tabref{tab:ablation_nuScene_table}, we find that the deep fusion module can slightly improve the results for all three baseline detectors. Interestingly, compared to the \textit{SECOND} and \textit{PointPillars}, \textit{CenterPoint} gives much better performance by adding the deep fusion module. This can be explained that the large deep backbone network in \textit{CenterPoint} gives much deeper features that are more suitable for the proposed deep feature fusion module. 

\begin{table}[t]
\centering
\renewcommand\arraystretch{1.1}
\setlength{\tabcolsep}{1pt}
\resizebox{0.5\textwidth}{!}
{%
\begin{tabular}{r|ccc|c} 
\hline
\multirow{2}{*}{\textbf{Strategies}} & \multicolumn{3}{c|}{\textbf{AP(\%)}}                                                                            & \multirow{2}{*}{\textbf{mAP(\%)}}        \\
                            & \multicolumn{1}{l}{Car(Mod.)}    & \multicolumn{1}{l}{Ped.(Mod.)}   & \multicolumn{1}{l|}{Cyc.(Mod.)}  &                                 \\ 
\hline
SECOND                      & ~88.99~                          & 56.21                            & 70.65                            & 71.95                           \\
One scale CA                & ~89.68 (\textcolor{red}{+0.69})~ & 57.77 (\textcolor{red}{+1.56})   & ~71.39 (\textcolor{red}{+0.74})~ & 73.70 (\textcolor{red}{+1.75})  \\
Multi-scale CA              & 90.22 (\textcolor{red}{+1.23})   & ~58.11 (\textcolor{red}{+1.90})~ & ~71.77 (\textcolor{red}{+1.12})~ & 74.41 (\textcolor{red}{+2.46})  \\
\hline
\end{tabular}
}
\caption{Ablation with/without multi-scale channel attention (CA) on KITTI dataset. The meaning of \textit{one-scale CA} is we just use $F_{R}^{L}$ to the \textit{CA} module. And multi-scale denote that we use both  fused $F_{R}^{L}$ and $F_{R}^{S}$ feature to the next module.}
\label{tab:ablation_channel_attention_kitti}
\end{table}

\noindent\textbf{Effectiveness of Multi-scale CA.} {In addition, a small experiment is also designed for testing the effectiveness of multi-scale attention in the DFF module. \textit{SECOND} is taken as the baseline and tested on the KITTI dataset. From the results given in Tab. \ref{tab:ablation_channel_attention_kitti}, we can see that by employing the one-scale features, the \textbf{mAP} has been improved by 1.75 points compared to the baseline. By adding the multi-scale operation, additional 0.71 points improvements can be further obtained.}

\begin{figure*}[h]
	\centering
	\includegraphics[width=0.975\textwidth,height=13.3cm]{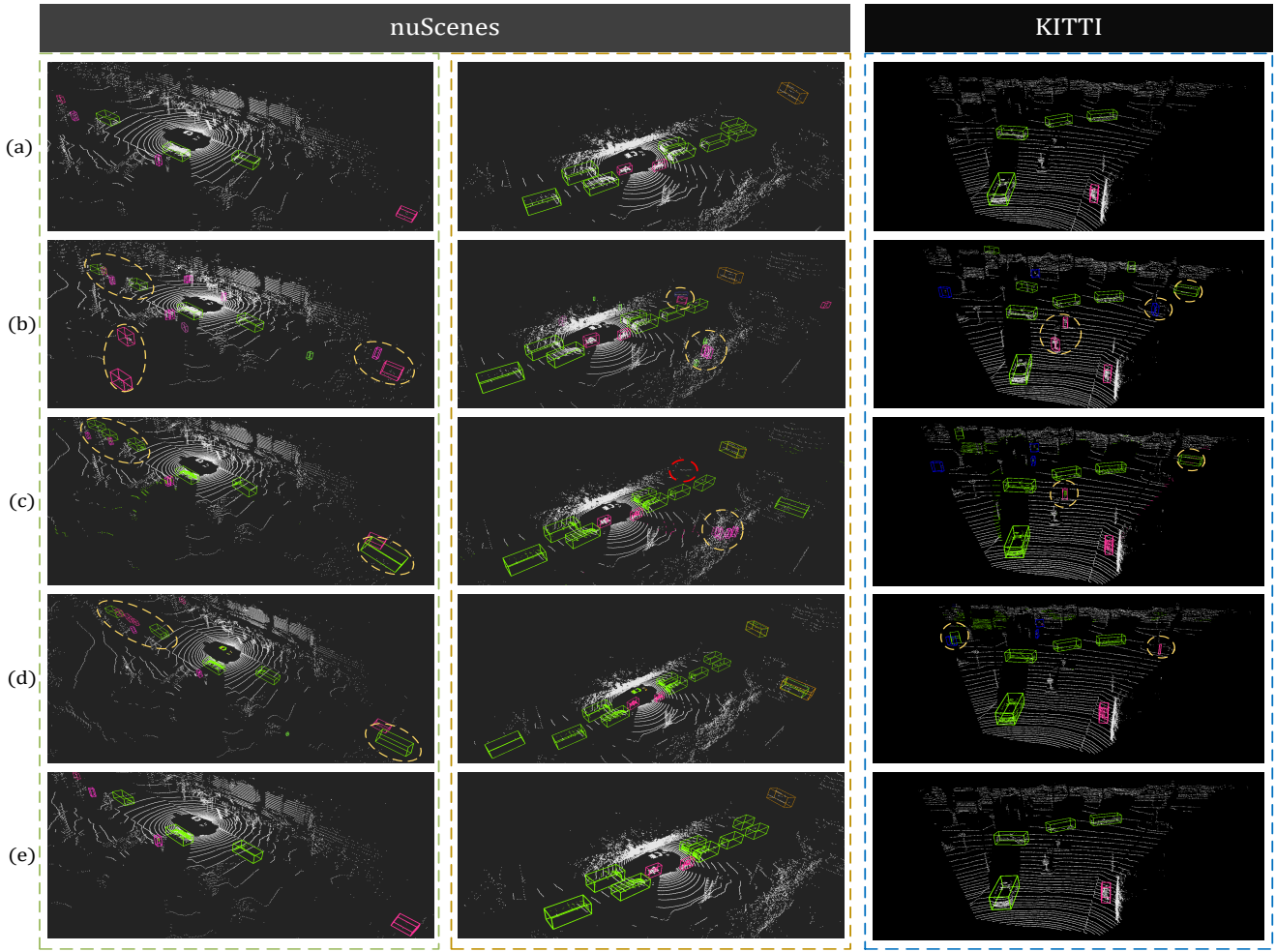}
	\centering
	\caption{Visualization of detection results with Open3d \cite{zhou2018open3d}, where (a) is the ground-truth, (b) is the baseline method based on only point cloud, (c), (d) and (e) are the detection results based on the 2D semantic, 3D semantic and fusion semantic information, respectively. Especially, the yellow and red dash ellipses show some false positive and false negative detection. For nuScenes dataset, the baseline detector used here is CenterPoint, and for KITTI is SECOND.} %
	\label{Fig:detection result}
\end{figure*}
\begin{table}[t]
\centering
\setlength{\tabcolsep}{5pt}
\renewcommand\arraystretch{1.1}
\resizebox{0.35\textwidth}{!}
{   
    \begin{tabular}{r | cccc}
    \hline
    $\textbf{Types}$ & $\textbf{PointPillars }$ & $\textbf{SECOND}$ \\ \hline 
    {Baseline}      & 53.5 (ms) & 108.2 (ms) \\
    {+3D Sem}       & +3.0 (ms) & +0.5 (ms)    \\ 
    {+2D Sem}       & +3.4 (ms) & +0.2 (ms)    \\ 
    {The Proposed}   & 62.5 (ms) & 113.3 (ms)   \\ \hline
    \end{tabular}
}
\caption{Inference time of different modules. Here, \textit{+3D Sem} and \textit{+2D Sem} denote the time of adding 3D/2D semantic information to input data. \textit{The Proposed} denotes the proposed framework with the modules including two types of semantic information and \textit{AAF} \& \textit{DFF} modules.}
\label{tab:inference_time}
\end{table}

{\subsection{Computation Time}} \label{subsec:comp_time}

Besides the performance, the computation time is also very important. Here, we test the computation time of each module based on the {\textit{PointPillars} and \textit{SECOND} on the KITTI dataset in \tabref{tab:inference_time}.} For a single Nvidia V100 GPU, the \textit{PointPillars} takes 53.5 ms per frame. By adding the 2D and 3D semantic information, the inference time increases 3.0 ms and 3.4 ms respectively while the time increases about 9 ms by adding two types semantic information and the AFF \& DFF modules. The inference time for \textit{SECOND} is given at the right column of \tabref{tab:inference_time}. Compared with the baseline, the 2D/3D semantic segmentation information gives nearly no extra computational burden. We explain this phenomenon as that in \textit{SECOND} the simple \textit{mean} operation is employed for extracting the features for each voxel and the computation time of this operation will not change too much with the increasing of the feature dimension. For \textit{PointPillars}, the MPL is employed for feature extracting in each pillar, therefore the computation time will increase largely with the increasing of the feature dimension.

{In addition, we also record the time used for obtaining the 2D/3D semantic results. For Deeplab V3+, the inference time is about 32 ms per frame while for Cylinder3D, it takes about 140 ms per frame. Furthermore, the re-projection of 2D image to 3D point clouds also takes about 3 ms for each frame. The almost time consuming here are 2D/3D point clouds segmentation operations. But in the practical using there just need a extra segmentation head after detection network backbone. In other words, multi-head is needed here for both detection and segmentation task. These just taking few milliseconds when using model inference acceleration operation, like C++ inference library TensorRT.}
\\

\subsection{Qualitative Detection Results}\label{sub:Qualitative_Res}
We show some qualitative detection results on nuScenes and KITTI dataset in \figref{Fig:detection result} in which \figref{Fig:detection result} (a) is the ground truth, (b), (c), and (d) are the detect result of baseline (CenterPoint) without any extra information, with 2D and 3D semantic information respectively and (e) is final results with all the fusion modules. From these figures, we can easily find that there is some false positive detection caused by the frustum blurring effect in 2D painting, while the 3D {semantic results} give a relatively clear boundary of the object but provides some worse class classification. More importantly, the proposed framework which combines both the two complementary information from 2D and 3D segmentation can give much more accurate detection results.

\section{Conclusion and Future Works}

{In this work, we proposed an effective framework \textit{Multi-Sem Fusion} to fuse the RGB image and LiDAR point clouds in two different levels. For the first level, the proposed AAF module aggregates the semantic information from both the 2D image and 3D point clouds segmentation, resulting in adaptivity with learned weight scores. For the second level, a DFF module is proposed to fuse further the boosted feature maps with different receptive fields by channel attention module. Thus, the features can cover objects of different sizes. More importantly, the proposed modules are detector independent, which can be seamlessly employed in different frameworks. The effectiveness of the proposed framework has been evaluated on public benchmark and outperforms the state-of-the-art approaches. However, the limitation of the current framework is also obvious. Both the 2D and 3D parsing results are obtained by offline approaches, which prevent the application of our approach in real-time AD scenarios. An interesting research direction is sharing the backbone features for object detection and segmentation and taking the segmentation as an auxiliary task.}

\ifCLASSOPTIONcaptionsoff
  \newpage
\fi

\bibliographystyle{ieeetr}
\bibliography{ref}







\begin{IEEEbiography}[{\includegraphics[width=1in,height=1.25in,clip,keepaspectratio]{{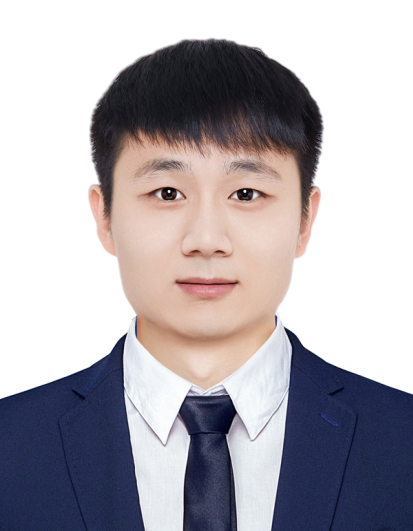}}}]{Shaoqing Xu} received his M.S. degree in transportation engineering from the School of Transportation Science
and Engineering in Beihang University. He is currently working toward the Ph.D. degree in electromechanical engineering with the State Key Laboratory of Internet of Things for Smart City, University of Macau, Macao SAR, China. His research interests include intelligent transportation systems, Robotics and computer vision.
\end{IEEEbiography}

\begin{IEEEbiography}[{\includegraphics[width=1in,height=1.25in,clip,keepaspectratio]{{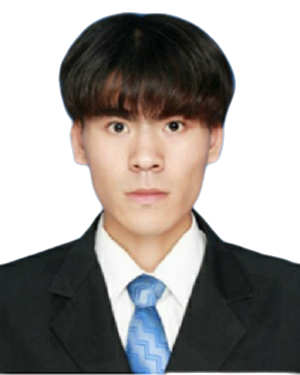}}}]{Fang Li} received the B.S. degree from Harbin Institute of Technology, Weihai, China, in 2020. His research focuses on LiDAR-based 3D object detection during the graduate study and received M.S. degree in mechanical engineering in Beijing Institute of Technology, Beijing, China. His research interests include 3D Object Detection and applications in Autonomous Driving. 
\end{IEEEbiography}

\begin{IEEEbiography}[{\includegraphics[width=1in,height=1.25in,clip,keepaspectratio]{{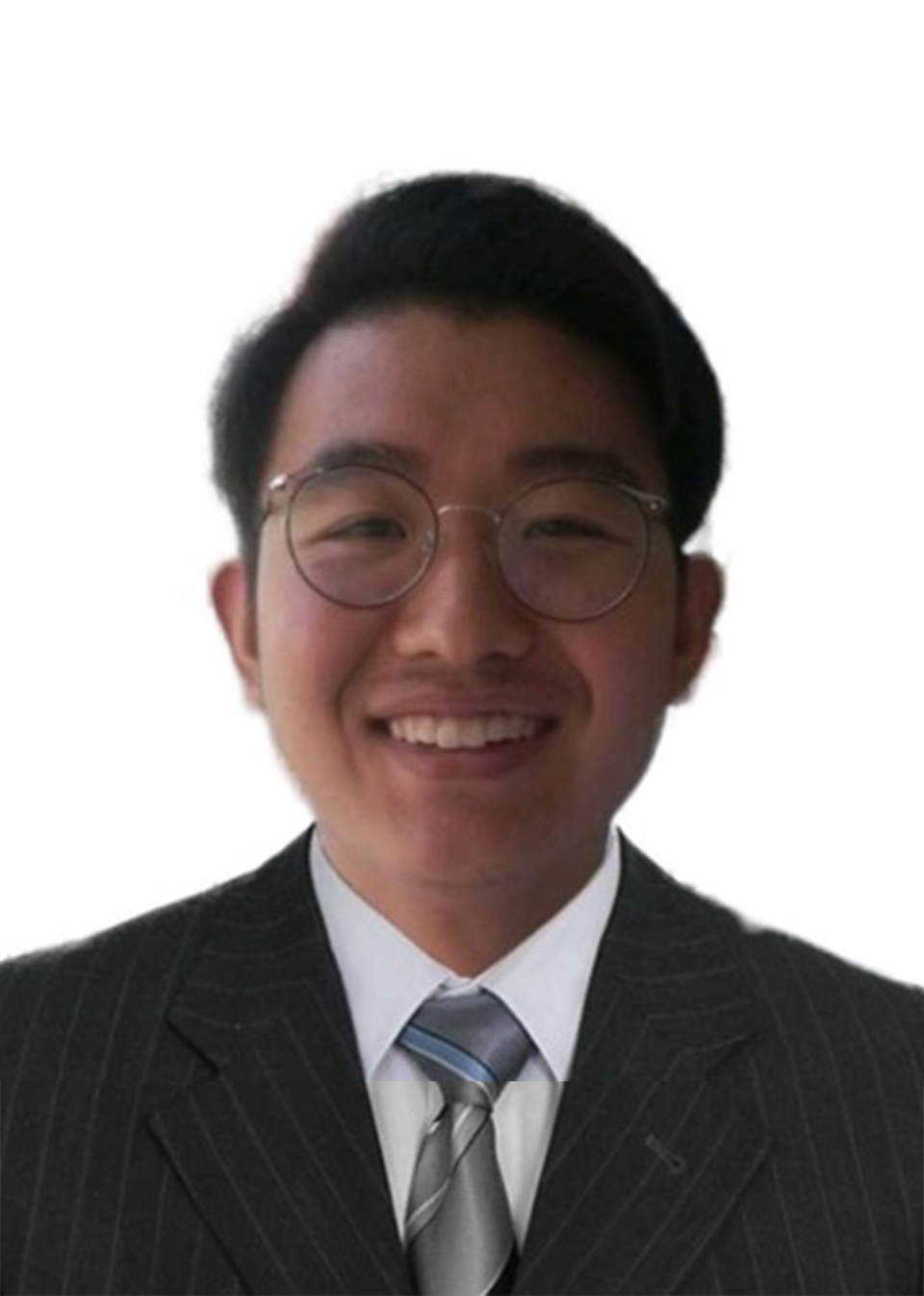}}}]{Ziying Song} was born in Xingtai, Hebei Province, China, in 1997. He received his B.S. degree from Hebei Normal University of Science and Technology (China) in 2019. He received a master's degree from Hebei University of Science and Technology (China) in 2022. He is now a Ph.D. student majoring in Computer Science and Technology at Beijing Jiaotong University (China), with research focus on Computer Vision. 
\end{IEEEbiography}

\begin{IEEEbiography}[{\includegraphics[width=1in,height=1.25in,clip,keepaspectratio]{{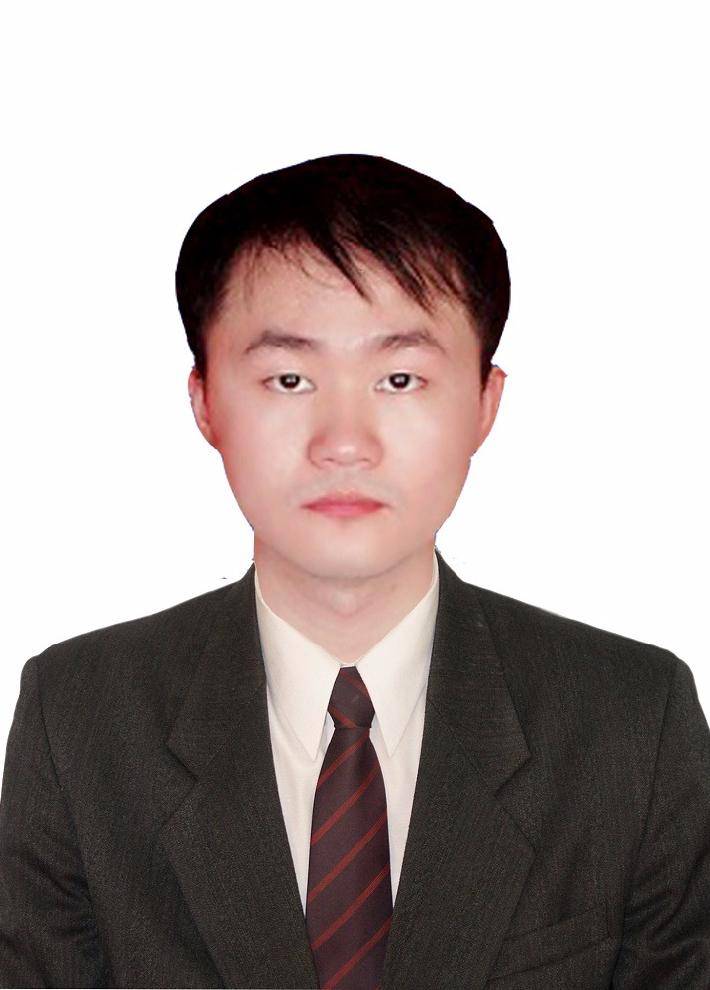}}}]{Jin Fang} is currently a researcher with Inceptio. He obtained his B.E. degree from Huazhong University of Science and Technology, and received the M.E degree in Information Engineering from Peking University, in 2016. Currently, his research interests include LiDAR Simulation, 3D Object Detection, and applications in Autonomous Driving.
\end{IEEEbiography}

\begin{IEEEbiography}[{\includegraphics[width=1in,height=1.25in,clip,keepaspectratio]{{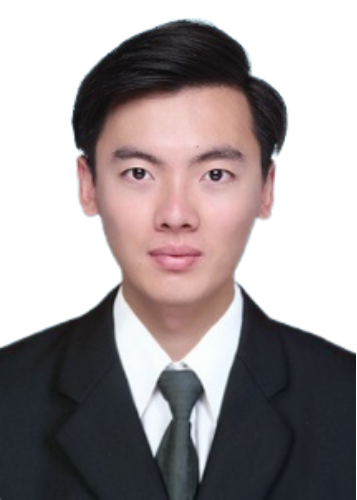}}}]{Sifen Wang} received his M.S. degree in transportation engineering in 2020 from Beihang University. He is currently working toward the Ph.D. degree in the School of Transportation Science and Engineering, Beihang University, Beijing, China. His research interests include intelligent vehicle, deep reinforcement learning and computer vision.
\end{IEEEbiography}

\begin{IEEEbiography}[{\includegraphics[width=1in,height=1.25in,clip,keepaspectratio]{{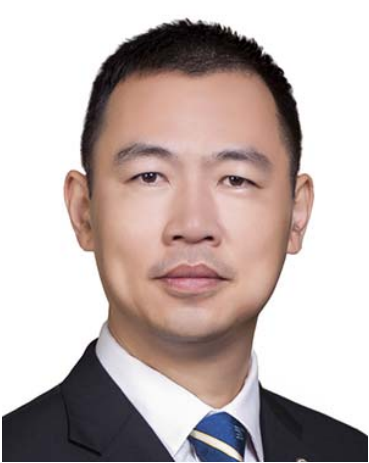}}}]{Zhi-Xin Yang} (Member, IEEE) received the
B.Eng. degree in mechanical engineering from the Huazhong University of Science and Technology, and the Ph.D. degree in industrial engineering and engineering management from the Hong Kong University of Science and Technology, respectively. He is currently an Associate Professor with the University of Macau. His current research interests include robotics, machine vision, intelligent fault diagnosis and safety monitoring.
\end{IEEEbiography}

\end{document}